\documentclass[letterpaper, 10 pt, conference]{ieeeconf}  

\IEEEoverridecommandlockouts                              

\usepackage{cite}
\usepackage{comment}
\usepackage{todonotes}
\usepackage{amsmath,amssymb,amsfonts}
\usepackage{booktabs}
\usepackage{graphicx}
\usepackage{textcomp}
\usepackage{xcolor}
\def\BibTeX{{\rm B\kern-.05em{\sc i\kern-.025em b}\kern-.08em
    T\kern-.1667em\lower.7ex\hbox{E}\kern-.125emX}}
    
\usepackage{paralist}
\let\labelindent\relax
\usepackage[shortlabels]{enumitem}

\usepackage[ruled]{algorithm}
\usepackage{etoolbox}
\usepackage[noend]{algpseudocode}
\usepackage{MnSymbol}
\usepackage{siunitx}
\makeatletter
\def\BState{\State\hskip-\ALG@thistlm}
\makeatother

\usepackage{microtype}
\usepackage{amsmath}
\usepackage{booktabs, makecell, tabularx}

\setcellgapes{2pt}
\newcolumntype{L}{>{\raggedright\arraybackslash}X}
\usepackage{caption}
\usepackage{changepage}
\title{\Large \bf
Robot Trajectory Adaptation to Optimise the Trade-off between Human Cognitive Ergonomics and Workplace Productivity in Collaborative Tasks
\vspace{-0.4cm}
}

\author{
Marta Lagomarsino$^{1,2}$,
Marta Lorenzini$^1$,
Elena De Momi$^2$,
Arash Ajoudani$^1$
\vspace{-0.4cm}
\thanks{$^{1}$ Human-Robot Interfaces and Physical Interaction, Istituto Italiano di Tecnologia, Via S.Quirico 19d, Genoa, Italy}
\thanks{$^{2}$ Department of Electronics, Information and Bioengineering, Politecnico di Milano, Via Giuseppe Colombo 40, Milan, Italy} 
\thanks{Corresponding author's email: {\tt\small marta.lagomarsino@iit.it}}
\thanks{This work was supported in part by the ERC-StG Ergo-Lean (Grant Agreement No.850932), in part by the European Union’s Horizon 2020 research and innovation programme SOPHIA (Grant Agreement No. 871237).}
}

\begin{document}

\maketitle
\thispagestyle{empty}
\pagestyle{empty}

\begin{abstract}
In hybrid industrial environments, workers' comfort and positive perception of safety are essential requirements for successful acceptance and usage of collaborative robots. 
This paper proposes a novel human-robot interaction framework in which the robot behaviour is adapted online according to the operator's cognitive workload and stress. The method exploits the generation of B-spline trajectories in the joint space and formulation of a multi-objective optimisation problem to online adjust the total execution time and smoothness of the robot trajectories. The former ensures human efficiency and productivity of the workplace, while the latter contributes to safeguarding the user's comfort and cognitive ergonomics. The performance of the proposed framework was evaluated in a typical industrial task. Results demonstrated its capability to enhance the productivity of the human-robot dyad while mitigating the cognitive workload induced in the worker. 
\end{abstract}

\vspace{-0.1cm}

\section{INTRODUCTION}
\label{sec:introduction}

The trend towards Industry 5.0 will bring humans and machines together to enable resource-efficient and user-centred manufacturing. 
This vision goes beyond mere efficiency and productivity and aims to place the well-being of workers at the centre of industrial processes \cite{Maddikunta2021}. To bring this into reality, collaborative robots (CoBots) have been increasingly adopted in industries to relieve their human counterparts from the physical effort (e.g. introduced by handling heavy loads \cite{brosque2020human}) and provide support in accomplishing hazardous operations (e.g. dealing with chemical material \cite{liu2020remote}).
Nevertheless, the coexistence of humans and Cobots in the same workplace may lead to adverse health outcomes in terms of mental stress and anxiety \cite{vanderMolene2020}, which should be profoundly studied. 

\begin{figure}[!t]
    \centering
    \includegraphics[width=0.95\linewidth]{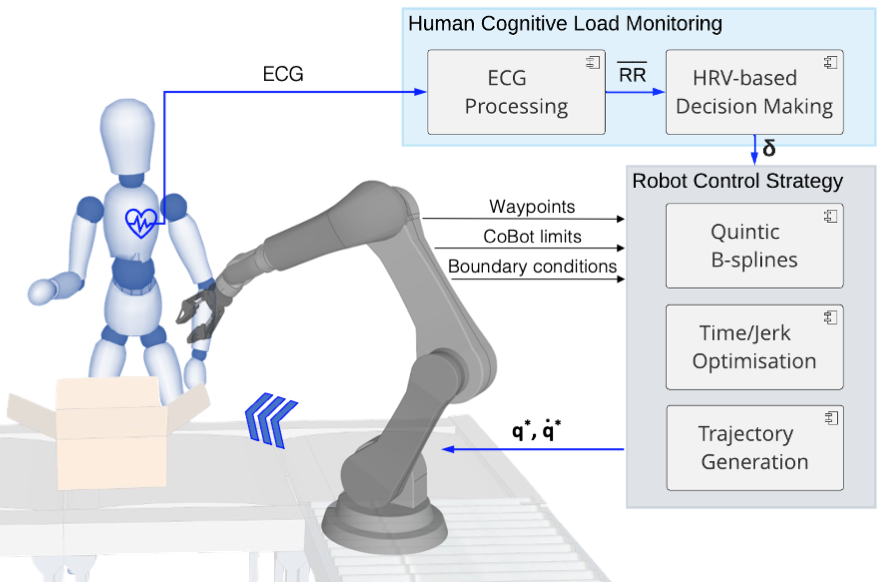}
    \caption{\small Overall procedure for psychologically safe HRC.}
    \label{fig:schema}
    \vspace{-0.7cm}
\end{figure}

In view of the prosperity of the next generation production lines, it is crucial to design human-robot collaboration (HRC) systems capable of online assessing the operator's mental fatigue 
and adapting the behaviour of the robotic teammate when needed. 
Research following this principle, usually referred to as ``affective robotics'' (i.e. the combination of robotics and affective computing), relies on the monitoring and interpretation of nonverbal communication, such as gaze direction\cite{Dini2017}, body language\cite{Lagomarsino2021, Lagomarsino2022} and physiological signals (i.e. cardiac \cite{Villani2018, Messeri2021} and electrodermal \cite{Hopko2021} activity).  
The requirement to continuously gather information about the mental processes is motivated by the high subjectivity and multidimensional construct of psychological status. Besides, due to the cumulative nature of the mental fatigue, the operator's needs may vary along with the task execution. 

A first attempt toward affective robotics in industrial settings was made in \cite{Villani2018}, where the remote control of a mobile robot was simplified when the user's cognitive load, estimated by analysing the heart rate variability (HRV), was exceeding human capabilities. 
Assistance to stressed users was provided by reducing the maximum velocity and assigning an almost autonomous behaviour to the robot. 
Besides, the work of \cite{Messeri2021} modelled HRC as a repeated non-cooperative game between two self-interested players, i.e. the human and the robot.
The former is assumed to aim at minimising his/her stress while the latter at maximising productivity. So, the state of collaboration was estimated, and CoBot pace adjusted accordingly. However, a great scope of improvement is envisaged on this topic addressing not only the robot timing but also its behaviour and adaptation capabilities. 

To contribute to tackling the ambitious challenge, this paper proposes a framework to simultaneously optimise the human cognitive workload and productivity during the collaboration by online adapting the CoBot's trajectory (see Fig. \ref{fig:schema}). 
A multi-objective optimisation problem is implemented to tune the total execution time and smoothness of the joint trajectories accomplished by the robot. The smoothness is achieved by generating trajectories featuring continuous joint acceleration, which yield a limited absolute value of the jerk (i.e. the third time derivative of position). Such smooth robot manoeuvres have shown to contribute to a safer and more comfortable sensation by the human co-workers \cite{Kuehnlenz2016, Kuehnlenz2018}, which might be due to their resemblance to natural arm movements \cite{Flash1985}.
Minimum-jerk trajectories have the potential to ensure cognitive ergonomics, reducing the mental demand \cite{Rojas2020}. However, the pure minimisation of the jerk will lead to slow robot motions, compromising the productivity of the human-robot dyad. Therefore, an objective to minimise the total time needed for following the path is included.
An online decision-making algorithm is then implemented to select the most appropriate solution, balancing a trade-off between robot execution speed and the user's mental fatigue, which is continuously monitored by analysing HRV. 

To the best of our knowledge, it is the first time that the CoBot's trajectory is adapted online to keep the human cognitive workload bounded and ensure psychologically safe HRC. The proposed framework was tested in a collaborative task, where the human and the CoBot were asked to work together and achieve a common goal. 

\section{
FAST and PSYCHOLOGICALLY SAFE TRAJECTORIES for COBOTS
}
\label{sec:methods}

The working principle of the psychologically safe HRC framework is highlighted in Figure \ref{fig:schema}. 
The goal is to solve the trade-off between system productivity and acceptable amount of human cognitive workload\footnotetext{Throughout the paper, ``cognitive load'' refers to the amount of mental processing and stress inducted by HRC. Note that the term ``stress'' is occasionally used as a synonym to simplify the argumentation.} in industrial tasks. 

To enhance the intuitiveness and flexibility of the framework, we implemented a waypoint-based trajectory planner. Indeed, the user can easily teach the CoBot a new task by manually positioning and orienting the joints in a sequence of desired poses, referred to as waypoints. 
A path passing through the waypoints is thus generated, whose smoothness and the associated timing law depend on the operator's psychological state. 
Proper timing laws can be used to generate fast CoBot trajectories, guaranteeing the productivity of the human-robot dyad. However, work intensification in the long term might induce a non-negligible cognitive load.  
Hence, the smoothness can be tuned to safeguard the user’s comfort and cognitive ergonomics. 

Our framework monitors the human cognitive load by analysing the heart rate variability and continuously regulates the robot joint trajectories. The purpose is to find the most appropriate pace of interaction for each specific user and online adapt it to fulfil changes in the individual needs. This allows keeping the human mental demand and stress bounded while maximising system productivity.  

\subsection{Quintic B-spline Trajectories in the Joint Space}
\label{sec:trajectories}

The trajectory planning problem exploits the formulation of B-splines in the joint space. 
A general B-spline curve of degree $p$ is defined as 
\vspace{-0.3cm}
\begin{equation}
\label{eq:bspline}
    b(t) = \sum_{k=1}^{C+1} c_k \, N_{k,p}(t),
\vspace{-0.1cm}
\end{equation}
where $N_{k,p}(t)$ are basis functions expressed recursively by the De Boor formula \cite{Boor2001} through the definition of a sequence of $M+1$ knots $[\tau_1, \tau_2, \dots \tau_{M\!+\!1}]$ in the interval $t \in [0,t_\text{f}]$
\begin{equation}
\vspace{-0.1cm}
\label{eq:deboor}
\begin{cases}
    N_{k,p}(t) = \frac{t-\tau_k}{\tau_{k+p}-\tau_k} N_{k,p-1}(t) + \frac{\tau_{k+p+1}-t}{\tau_{k+p+1}-\tau_{k+1}} N_{k+1,p-1}(t) \\
    N_{k,0}(t) = 
    \begin{cases}
        1, & \text{if } \tau_k\leq t <\tau_{k+1} \\
        0, & \text{otherwise.}
    \end{cases}
    \vspace{-0.1cm}
\end{cases}
\end{equation}
The $C\!+\!1$ coefficients $c_k$ of the linear combination are named control points and define a polygon in which the curve $b(t)$ is contained. Another key feature of this curve is that its derivatives are continuous up to the $\{p\!-\!1\}$-th derivative, i.e. $\mathcal{C}^{p-1}$. For further details about B-spline curves, the reader is referred to \cite{Gasparetto2007}. 

Within the proposed control strategy, we generate B-spline curves of degree five to obtain trajectories with continuous forth-derivative of the joint position, thus guaranteeing bounded values of joint velocity, acceleration and jerk along the path.
Moreover, $p\!=\!5$ permits imposing boundary conditions till the acceleration. 
Each joint trajectory is subject to the condition of passing through $W$ waypoints $\{w_1, w_2, \dots, w_W\}^j$ at a sequence of unknown time instants $\{t_1, t_2,\dots, t_W\}$. 
In the next section, the procedure to find the optimal time intervals between consecutive waypoints is defined.

\begin{figure}[!b]
    \centering
    \vspace{-0.4cm}
    \includegraphics[width=\linewidth]{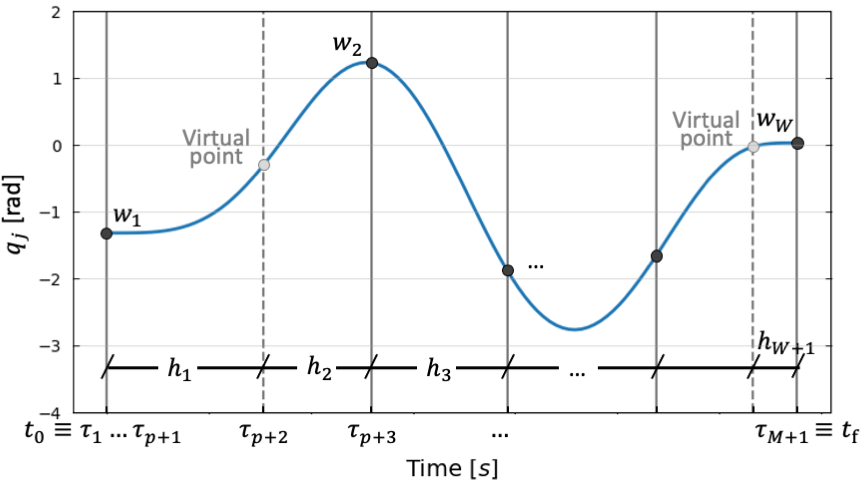}
    \caption{\small B-spline curve passing through a set of waypoints (black dots) and two virtual points (gray dots). Time intervals $h_l$ between consecutive points, defining optimisation vector $\bf{h}$, are highlighted.}
    \label{fig:bspline}
\end{figure}

Given the intervals vector, the knot vector $\tau$ can be filled. 
The values at the extremities feature multiplicity $p\!+\!1$, so that the first and last control points coincide with the desired waypoints $w_1$ and $w_W$ and are attained at $t_1\!=\!0$ and $t_W\!=\!t_\text{f}$. 
Two virtual points have been additionally introduced to impose zero boundary conditions also for the jerk in fifth-degree curves. 
As a result, each joint $j$ position trajectory is defined as
$ q_j(t) = b(t)$ with $p=5$, $\tau$ is composed by $M\!+\!1=(W\!+\!2)+2p=W\!+\!12\,$ knots (see Fig. \ref{fig:bspline}) and 
the number of required control points $c_{k,0}^j$ is $C\!+\!1=W\!+\!p\!+\!1=W\!+\!6$. 

The derivative of a B-spline curve of degree $p$ is a B-spline of degree $p\!-\!1$. Therefore, joint velocity, acceleration and jerk result in 
\vspace{-0.4cm}
\begin{align}
    \dot{q}_j(t) =& \sum_{k=1}^{C} c_{k,1}^j \, N_{k,p-1}(t), \label{eq:velocity}\\
    \ddot{q}_j(t) =& \sum_{k=1}^{C-1} c_{k,2}^j \, N_{k,p-2}(t), \\
    \dddot{q}_j(t) =& \sum_{k=1}^{C-2} c_{k,3}^j \, N_{k,p-3}(t).
\end{align}
where the control points of $d$-th derivative are expressed as 
\begin{equation}
    c_{k,d}^j = 
    \begin{cases}
        c_{k,0}^j & \text{if } d=0\\
        (p+1-d)\dfrac{c_{k,d-1}^j-c_{k-1,d-1}^j}{\tau_{k+p+1-d}-\tau_k} & \text{otherwise.}
    \end{cases}
\end{equation}

\subsection{Multi-objective Time/Jerk Trajectory Optimisation}
\label{sec:optimisation}
 
The robot trajectory based on B-splines in the joint space is optimised by simultaneously minimising the total execution time and the integral of squared jerk (i.e. maximise the smoothness) along the trajectory. 
It is trivial to notice that the two components have opposite effects: lessening the former results in rapidly completed trajectories that features large kinematic quantities values, while reducing the latter leads to more smooth trajectories taking more time to reach the target pose. 
Ultimately, we aim to find the optimal trade-off in the current human cognitive load condition. 

A multi-objective optimisation problem is designed as
\begin{equation}
\begin{aligned}
    \underset{\textbf{h}}{\text{min}}  \quad & f_\text{time}(\textbf{h})=D\,\sum_{l=1}^{W+1}{h_l} \\
    \underset{\textbf{h}}{\text{min}}  \quad & f_\text{jerk}(\textbf{h})=\sum_{j=1}^{D}{\int_{0}^{t_\text{f}} \left(\sum_{k=1}^{C-2} c_{k,3}^j \, N_{k,p-3}(t)\right)^2 dt} 
    \\
    s.t. \quad & |\dot{q}_j(t)|\leq v^{\text{max}}_j \\
    \quad & |\ddot{q}_j(t)|\leq a^{\text{max}}_j \quad\quad\quad\quad j=1,\dots,D \\
    \quad & |\dddot{q}_j(t)|\leq \psi^{\text{max}}_j\\
    \quad & h_l\geq h_l^{\text{lb}} \ \quad\quad\quad\quad l=1,\dots,W+1 \\ 
\end{aligned}
\end{equation}
where the variables that should be optimised\footnotemark  are the time intervals between consecutive waypoints (see Fig.\ref{fig:bspline}), 
namely $h_l=\tau_{p+l+1}-\tau_{p+l}$, and are grouped in the decision vector 
\vspace{-0.1cm}
\begin{equation}
    \textbf{h}=[h_1, h_2,\dots, h_{W+1}].
    \vspace{-0.1cm}
\end{equation}
\footnotetext{Note that the basis function $N_{k,p-3}$ can be expressed in analytical form as a function of \textbf{h} through the recursive formula in (\ref{eq:deboor}). This allows us to compute the integral of the squared jerk in $f_{\text{jerk}}$. However, the computation is extremely verbose, thus a numerical integration procedure can also be exploited at the cost of increasing the computation time.}
\vspace{-0.3cm}

The optimisation is subject to inequality constraints on the joints velocity, acceleration and jerk.
Exploiting the convex hull property of B-splines, those constraints can be expressed in terms of the corresponding control points, $c_{k,1}^j$, $c_{k,2}^j$,  $c_{k,3}^j$:
\vspace{-0.1cm}
\begin{equation}
\begin{aligned}
    |c_{k,1}^j|= & 
    \bigg|\frac{p(c_{k+1,0}^j-c_{k,0}^j)}{\sum_{z=\text{max}(1,k-p+1)}^k h_z}\bigg| 
    \leq v^{\text{max}}_j
    \, & k=1,..C \\
    |c_{k,2}^j|= & 
    \bigg|\frac{(p-1)(c_{k+1,1}^j-c_{k,1}^j)}{\sum_{z=\text{max}(1,k-p+2)}^k h_z}\bigg| 
    \leq a^{\text{max}}_j
    \, & k=1,..C-1 \\ 
    |c_{k,3}^j|= & 
    \bigg|\frac{(p-2)(c_{k+1,2}^j-c_{k,2}^j)}{\sum_{z=\text{max}(1,k-p+3)}^k h_z}\bigg| 
    \leq \psi^{\text{max}}_j \, & k=1,..C-2. \\
\end{aligned}
\end{equation}
\noindent Notice that this is a sufficient condition for the validity of the constraints.
Besides, optimisation variables are lower bounded 
\begin{equation}
    h_l^{\text{lb}}=\max_{\substack{j=1,\dots D}} \bigg\{\frac{|w_{l+1}^j-w_{l}^j|}{v^{\text{max}}_j}\bigg\},
\end{equation}
since any interval between a pair of consecutive waypoints ($w_i^j, \, w_{i+1}^j$) cannot be run at infinite velocity. 

\begin{figure}[!b]
    \vspace{-0.3cm}
    \centering
    \includegraphics[width=0.93\linewidth]{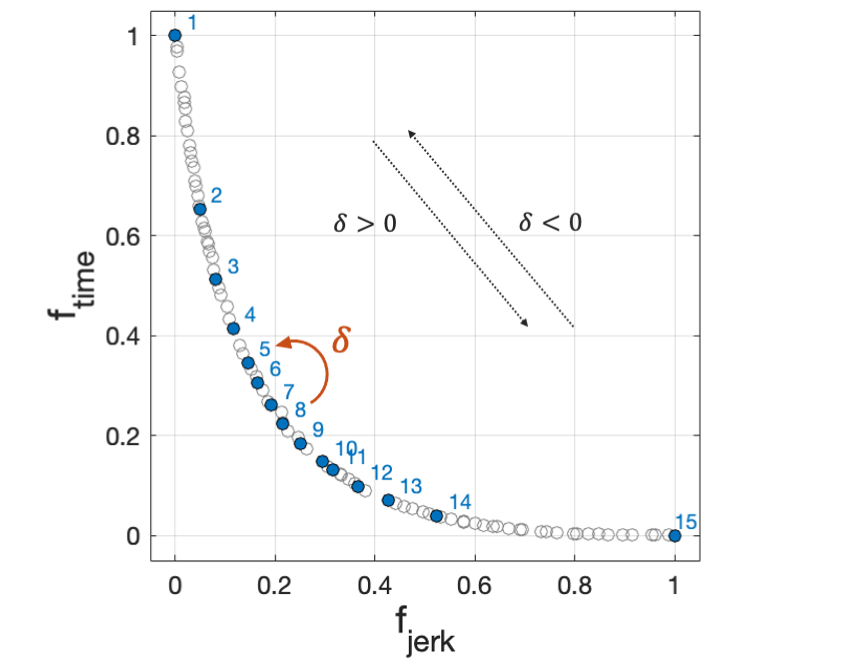}
    
    \vspace{-0.2cm}
    \caption{\small Pareto Optimal Front $\mathcal{F}$ (gray circles, normalised for sake of clarity) resulting by minimum time-jerk optimisation. Blue dots denote the front $\mathcal{F}^*$ after the downsampling procedure.}
    \label{fig:pareto}
\end{figure}

The constrained bi-objective problem is addressed using the well-known multi-objective algorithm NSGA-II \cite{Deb2002}. 
This procedure consists of computing a set of non-dominated solutions in the objective space, namely solutions for which none of the objective functions can be improved in value without deteriorating the other objective.
This set is called Pareto optimal front $\mathcal{F}$ and reveals the essential trade-offs between the objectives $f_\text{time}$ and $f_\text{jerk}$, represented with gray circles in Fig.\ref{fig:pareto}.
Each solution in $\mathcal{F}$ corresponds to a diverse interval vector \textbf{h} in the parameter space, thus to robot trajectories with different features.  
Throughout the paper, the resulting set of interval vectors will be referred to as $\mathcal{X}$. 

\subsection{Online Human Cognitive Load Monitoring}
\label{sec:HRV}

An online decision-making algorithm is implemented to select the most appropriate interval vector $\bf{h}$ from the set $\mathcal{X}$ according to the human socio-physical workload. 
The latter is continuously monitored during the interaction by analysing the heart rate variability (HRV). 

HRV has been identified as a highly sensitive marker of physiological conditions that likely alters the balance between sympathetic and parasympathetic nervous systems.  
The analysis of HRV relies on the computation of the time intervals (also referred to as RR intervals) between two consecutive heartbeats (i.e. R peaks) and their variation over time. 
Among various metrics proposed in the literature, the mean RR interval ($\overline{\text{RR}}$) was selected since it appeared to be the most sensitive measure of the mental effort 
and, in contrast to other measures, was proved to be reliable also when calculated in ultra-short recordings ($10$-$30$\si{\second}) \cite{Nussinovitch2011}. 

In this work, we periodically compute the mean value $\overline{\text{RR}}_i$ within a window of fixed time duration (at iteration $i$) and we compare it with the value at the previous iteration $\overline{\text{RR}}_{i-1}$. 
An intense cognitive demand leads to a decrease of $\overline{\text{RR}}$. 
When the variation exceeding a minimum value $\Delta_{r\rightarrow s}$ and $\overline{\text{RR}}_i$ is below the bound of the mean value at rest $\overline{\text{RR}}_r$, i.e. 
\vspace{-0.1cm}
\begin{equation}
\label{eq:stress}
    (\,\overline{\text{RR}}_i -\overline{\text{RR}}_{i-1} <-\Delta_{r\rightarrow s}\,) \wedge  (\,\overline{\text{RR}}_i<\overline{\text{RR}}_r\,),
\vspace{-0.1cm}
\end{equation}
a change toward higher mental workload level is detected. 
In this condition, we want to choose the interval vector out of the set $\mathcal{X}$ that boosts the smoothness of the desired robot trajectory at the cost of losing performance (i.e. assigning more importance to minimising the jerk than the total execution time). To this aim, we move in the objective space solutions $\mathcal{F}$ of a number of steps 
\vspace{-0.1cm}
\begin{equation}
\label{eq:step2stress}
    \delta = \Bigl\lfloor\dfrac{\Delta{\overline{\text{RR}}_r}+\Delta_{r\rightarrow s}}{\Delta_{r\rightarrow s}}\Bigr\rfloor <0,
\vspace{-0.05cm}
\end{equation} 
proportional to the variation of $\overline{\text{RR}}$ between consecutive time windows
\vspace{-0.4cm}
\begin{equation}
    \Delta{\overline{\text{RR}}_r} = \overline{\text{RR}}_i -\min(\overline{\text{RR}}_{r}, \, \overline{\text{RR}}_{i-1}).
\vspace{-0.1cm}
\end{equation}
Note that if $\overline{\text{RR}}_{i-1}$ is above the value at rest $\overline{\text{RR}}_r$, the variation $\Delta{\overline{\text{RR}}_r}$ is performed with respect to resting conditions. Also, the floor function in (\ref{eq:step2stress}) ensures that a slight variation in the stress range is mapped into a step in the solutions, while $N$ steps are computed with $\Delta{\overline{\text{RR}}_r}=-(N\!+\!1)\Delta_{r\rightarrow s}$. 

\begin{figure}[!t]
    \centering
    \vspace{-0.3cm}
    \includegraphics[width=\linewidth]{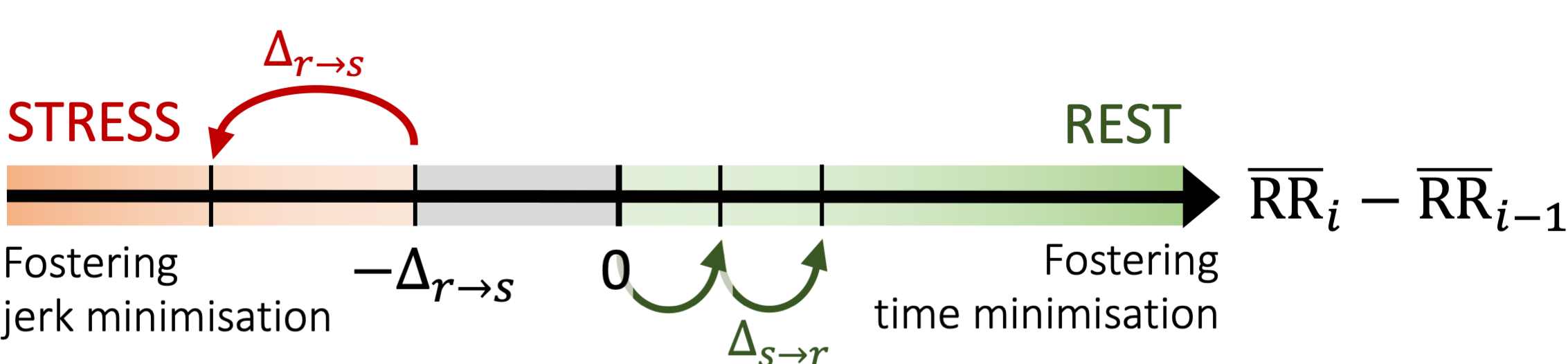}
    
    \vspace{-0.05cm}
    \caption{\small Detection of changes in cognitive load level in relation to $\overline{\text{RR}}_i$ variations.}
    \vspace{-0.4cm}
    \label{fig:rr_variations}
\end{figure}

Since our final aim is to enhance productivity without perilously increasing workers' cognitive demand, positive $\overline{\text{RR}}$ variations out of the range of stress (see Fig.\ref{fig:rr_variations}), i.e. 
\vspace{-0.1cm}
\begin{equation}
    (\, \overline{\text{RR}}_i -\overline{\text{RR}}_{i-1}>0) \wedge  (\,\overline{\text{RR}}_i\geq\overline{\text{RR}}_s\,),
\vspace{-0.1cm}
\end{equation}
are mapped into steps in the Pareto optimal front $\mathcal{F}$ toward faster but less smooth trajectories. In particular, $\delta$ is defined as 
\vspace{-0.2cm}
\begin{equation}
    \label{eq:step2rest}
    \delta = \Bigl\lceil\dfrac{\Delta{\overline{\text{RR}}_s}}{\Delta_{s\rightarrow r}}\Bigr\rceil >0
\end{equation} 
where the variation of $\overline{\text{RR}}$ is here computed in relation to stressful conditions, i.e. 
\vspace{-0.1cm}
\begin{equation}
    \Delta{\overline{\text{RR}}_s} = \overline{\text{RR}}_i -\max(\overline{\text{RR}}_{s}, \, \overline{\text{RR}}_{i-1}),
\vspace{-0.1cm}
\end{equation}
to avoid pushing productivity when the subject is already stressed.
Note that $\Delta_{s\rightarrow r}>\Delta_{r\rightarrow s}$, reflecting the hysteresis of the human body to switch between different mental states \cite{Villani2018}. This means that the minimum variation required to detect a change in the cognitive load level from rest to stress $\Delta_{r\rightarrow s}$ is higher than the transition between stress to rest $\Delta_{s\rightarrow r}$. Thus, we foster the minimisation of the execution time resulting in trajectories that features large values of kinematic quantities, though lower than the upper bounds imposed in the optimisation inequality constraints.
Moreover, the ceil function in  (\ref{eq:step2rest}) settles on one step toward faster trajectories when small positive variations $\Delta{\overline{\text{RR}}_s}$ are registered. More steps are performed every multiple of $\Delta_{s\rightarrow r}$.

Finally, no action is taken ($\delta\!=\!0$) with variations of $\overline{\text{RR}}$ in $[-\!\Delta_{r\rightarrow s}, 0]$. To prevent cumulative stress, it is checked that $\overline{\text{RR}}_i$ has not gradually reached undesirable low values during the task. On that occasion, a step toward more relaxing trajectories is performed ($\delta\!=\!-1$). 
The reader can find the main steps of our online decision-making algorithm in Alg.\ref{alg:CLmonitoring}.

\begin{algorithm}[!t]
\caption{HRV-based Decision Making}
\label{alg:CLmonitoring}
\begin{algorithmic}[1]
\Procedure{\textproc{DecisionMaker}}{}
\State $\Delta_{r\rightarrow s} \leftarrow$ select\_min\_$\overline{\text{RR}}$variation\_\textit{r}$\rightarrow$\textit{s}
\State $\Delta_{s\rightarrow r} \leftarrow$ select\_min\_$\overline{\text{RR}}$variation\_\textit{s}$\rightarrow$\textit{r}
\For {\textbf{each} path $\in \mathcal P$}
    \State $\mathcal{X}_{\text{path}} \leftarrow$ compute\_Pareto\_solution\_set(path)
    \State $k_{\text{path}} \leftarrow$ set\_initial\_solution\_index
\EndFor
\BState \emph{top}:
\For {\textbf{each} path $\in \mathcal P$}
    \State $\textbf{h}_{\text{path}} = \mathcal{X}_{\text{path}}[k_{\text{path}}, :]$
    \small\Comment{\emph{Select optimal solution}} \normalsize
\EndFor
\BState \emph{loop}: 
    \State $\overline{\text{RR}}_i \leftarrow$ compute\_mean\_RR
    \State $\Delta{\overline{\text{RR}}_i}\leftarrow \overline{\text{RR}}_i-\overline{\text{RR}}_{i-1}$;
    \If{$(\Delta{\overline{\text{RR}}_i} <-\Delta_{r\rightarrow s}) \wedge  (\overline{\text{RR}}_i<\overline{\text{RR}}_r)$}
        \State $\delta \leftarrow$ using \eqref{eq:step2stress}
        \small \Comment{\emph{Stress detected: fostering $f_{\text{jerk}}$}} \normalsize
    \ElsIf{$(\Delta{\overline{\text{RR}}_i}>0) \wedge  (\overline{\text{RR}}_i\geq\overline{\text{RR}}_s)$}
        \State $\delta \leftarrow$ using \eqref{eq:step2rest}
        \small \Comment{\emph{Rest detected: fostering $f_{\text{time}}$}} \normalsize 
    \ElsIf{$\overline{\text{RR}}_i<\overline{\text{RR}}_s^\text{max}$}
        \State $\delta \leftarrow -1$
        \small \Comment{\emph{Cumulative stress detected}} \normalsize
    \Else
        \State $\delta \leftarrow 0$
    \EndIf
    \State $k_{\text{path}} \leftarrow k_{\text{path}}+\delta$ 
    \State $i$++
\State \textbf{goto} \emph{top}.
\EndProcedure
\end{algorithmic}
\end{algorithm}

\subsection{Psychologically Safe Trajectories Generation}
\label{sec:trajectory_generation}

Once the optimal \textbf{h} is selected, the interpolation problem for each joint $j=1,..D$ can be solved by defining a system of $6+W$ equations. 
The six boundary conditions, namely initial and final values assumed by $\dot{\bf{q}}$, $\ddot{\bf{q}}$, and $\dddot{\bf{q}}$, are easily imposed by equalling them to the corresponding first and last control points (e.g. $c_{0,1}=\dot{\bf{q}}(0)=$ initial velocity)\footnote{Note that $\bf{q}$ refers to the configuration vector, i.e. vector of joints positions. Accordingly, $\dot{\bf{q}}$, $\ddot{\bf{q}}$, $\dddot{\bf{q}}$ are joint velocity, acceleration and jerk vectors.}.
The remaining $W$ equations express the passage of the joint trajectory ${\bf{q}}(t)$ through the waypoints (e.g. ${\bf{q}}(\tau_{(p+1)+3})=\bf{w}_3$). 
Since all equations can be expressed as a linear combination of the control points $c_{k,0}^j$ defining the joint position, the system can be rewritten as the linear system
\vspace{-0.1cm}
\begin{equation}
\label{eq:lin_system}
    {\bf{A}}\,{\bf{\Theta}}_j = {\bf{B}}_j,
\end{equation}
where the matrix \textbf{A} depends only on the interval vector \textbf{h} and is common to all joints, while the unknowns are the set of control points 
\vspace{-0.2cm}
\begin{equation}
    {\bf{\Theta}}_j=
    \begin{bmatrix}
        c_{1,0}^j\\
        \dots \\
        c_{C+1,0}^j
    \end{bmatrix}.
\vspace{-0.1cm}   
\end{equation}

Given the optimal \textbf{h} computed by the time-jerk optimisation and decision-making procedure (described in previous sections), the matrix \textbf{A} is computed and used in (\ref{eq:lin_system}) to obtain the control points for each joint $j$. As a result, the optimal trajectories for robot joints are fully defined: the desired joints position and velocity are computed by (\ref{eq:bspline}) and (\ref{eq:velocity}), respectively, and sent to the robot.

\section{EXPERIMENTS}
\label{sec:experiments}

The proposed control strategy was validated in a collaborative task, where the human and the robot were required to fill boxes before the dispatch. 
The task exemplified a scenario in which the robot has to perform a repetitive activity, but the human intervention enables the product customisation to fulfil task requirements and customer demand. 

\subsection{Task description}
\begin{figure}[!t]
    \centering
    \includegraphics[width=0.84\linewidth]{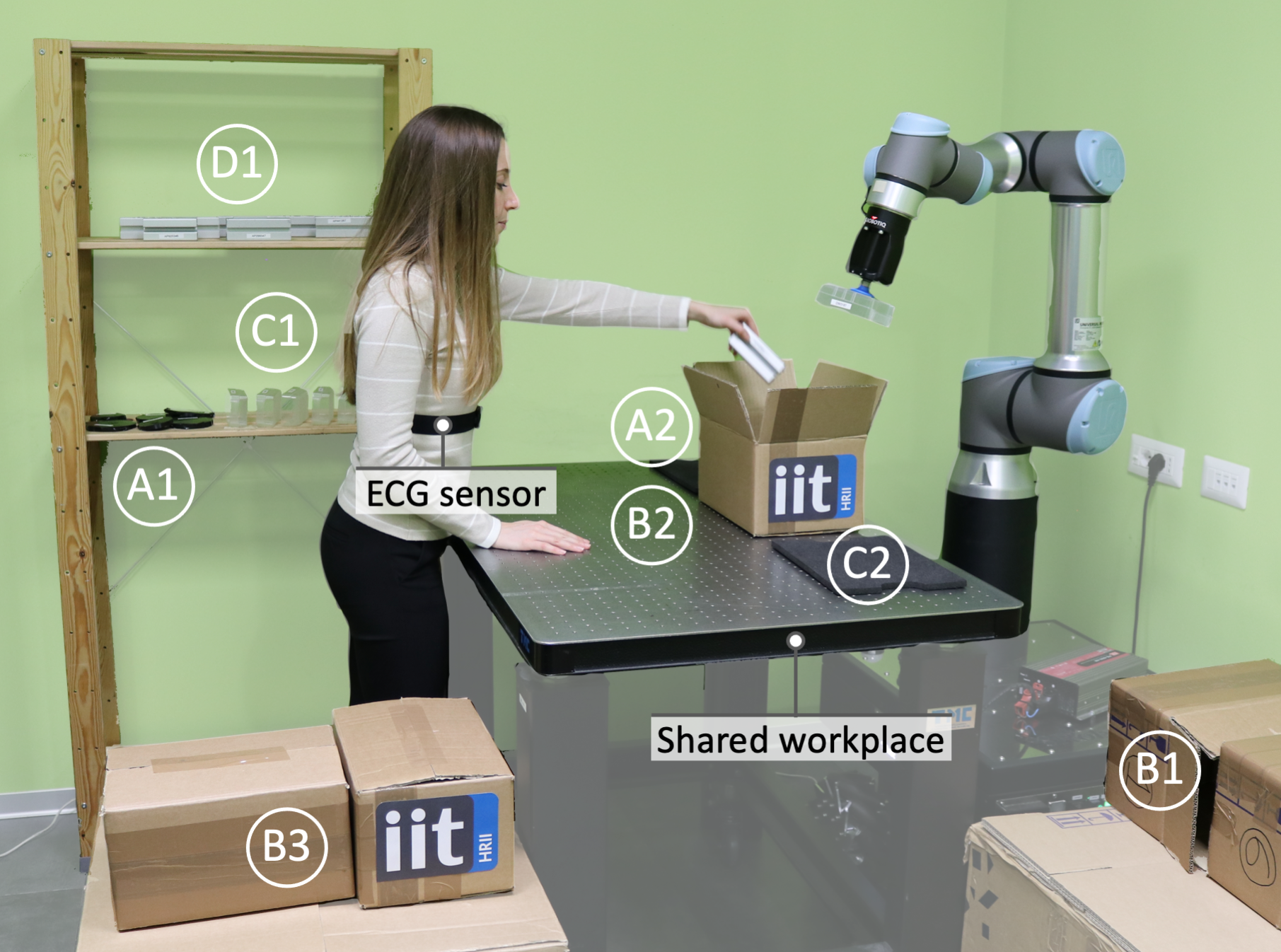}
    \caption{\small Overview of the experimental setup involving: human operator, industrial CoBot and shared workplace.}
    \vspace{-0.6cm}
    \label{fig:setup}
\end{figure}

The human-robot dyad was asked to performed cyclically the following collaborative task (experimental setup in Fig.\ref{fig:setup}). 
The industrial CoBot UR16e (Universal Robots, controlled at 500 Hz) equipped with the Robotiq’s vacuum gripper EPick picked a new empty box from \textit{B1} and brings it in \textit{B2}. Inside the box, there were three codes corresponding to objects that were present on a shelf. The participant was asked to select the correct objects from positions \textit{A1} and \textit{C1} and place them in \textit{A2} and \textit{C2}, respectively, located in the human-robot shared area. The CoBot, in turn, took them from the shared workplace and inserted them into the box. 
The third code matched an aluminium profile in location \textit{D1} that had to be put directly inside the box by the participant. Finally, the human closed the box and left it in position \textit{B3}. Meanwhile, the CoBot had grasped a new box, and a new cycle began. 

The task was designed to induce a mild cognitive load, as it occurs in real-world industrial environments. 
The electrocardiogram (ECG) signal of the participant was monitored by Polar H10 chest strap (130 Hz) for the entire duration of the experiment. Cardiovascular data were streamed to the CPU running the proposed framework via Bluetooth and processed to extract RR intervals. Every 30 seconds, the mean value of inter-beats intervals $\overline{\text{RR}}_i$ in the past window was computed and used to enhance the collaboration. The architecture and communications were implemented exploiting Robot Operating System (ROS) platform. 

\subsection{Experimental protocol}

Twelve healthy volunteers, seven males and five females ($26.9\pm1.7$ years old), were recruited within students and research personeel of Istituto Italiano di Tecnologia\footnote{The experimental protocol was approved by the ethics committee ASL Genovese N.3 (Protocol IIT\_HRII\_ERGOLEAN 156/2020).}.
Before the beginning of the experiment, the person-specific baseline values of ECG were recorded and the mean inter-beats interval in resting conditions $\overline{\text{RR}}_r$ was registered.

The CoBot tracked trajectories computed by interpolating a sequence of waypoints, defined a priori by the authors, with the method described above. 
The values of joints kinematic quantities at the extremities (i.e. boundary conditions) were fixed to zero to allow the pick and place of involved components, while the upper bounds $v^{\text{max}}_j$, $a^{\text{max}}_j$, and $\psi^{\text{max}}_j$ were set following UR16e specifications. The constrained bi-objective optimisation problem was solved using the NSGA-II algorithm with a population size of 90. The resulting Pareto front $\mathcal{F}$ and the corresponding solution set $\mathcal{X}$ was then downsampled (a possible solution in Fig.\ref{fig:pareto}).

\begin{figure}[!b]
    \centering
    \vspace{-0.6cm}
    \includegraphics[width=\linewidth]{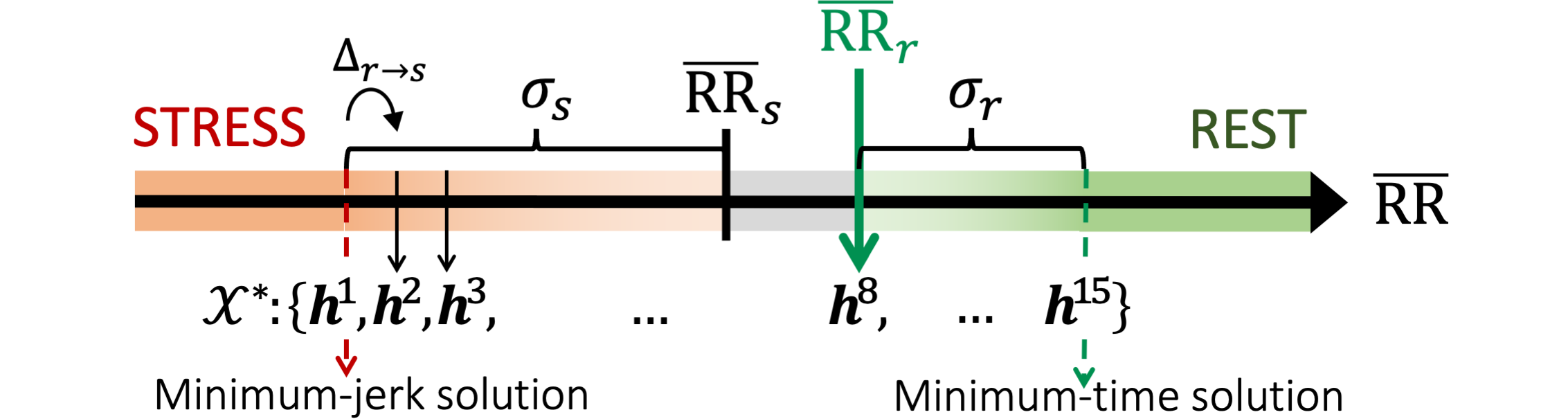}
    
    \vspace{-0.2cm}
    \caption{\small Initial mapping between $\overline{\text{RR}}$ intervals extracted by ECG signal and downsampled optimal solution set $\mathcal{X}^*$.}
    \label{fig:initial_mapping}
\end{figure}

To define the final samples and the parameters of the HRV-based decision-making algorithm, we referred to \cite{Villani2018}. Accordingly, we set $\Delta_{r\rightarrow s}=0.02$\si{\second} and $\Delta_{s\rightarrow r}=0.01$\si{\second}. 
In addition, we assumed that $\overline{\text{RR}}$ could span in the range $[\overline{\text{RR}}_s\!-\!\sigma_s, \, \overline{\text{RR}}_r\!+\!\sigma_r]$ (see Fig.\ref{fig:initial_mapping}), where $\sigma_s$ and $\sigma_r$ are the standard deviations registered by \cite{Villani2018} in stress and rest conditions. 
Thus, $\overline{\text{RR}}_s^\text{max}\!\!\!=\!\overline{\text{RR}}_s\!-\sigma_s$ and we determined the final number of solutions in the downsampled set $\mathcal{X}^*$ as $\frac{(\overline{\text{RR}}_r+\sigma_r)-(\overline{\text{RR}}_s-\sigma_s)}{\Delta_{r\rightarrow s}}\!\approx\!15$. 
The downsampling procedure was then implemented exploiting the decomposition method called Augmented Scalarization Function and computing the fifteen solutions by recursively assigning slightly more importance to $f_\text{time}$ with respect to $f_\text{jerk}$. 
As a result, the former element of $\mathcal{X}^*$ was the interval vector $\bf{h}^1$ corresponding to a minimum-jerk trajectory, while the latter, i.e. $\bf{h}^{15}$, defined a minimum-time trajectory. The remaining solutions expressed a trade-off between the two objectives. The higher the number of the selected solution, the less was the execution time $\gamma$ of a cycle.

The initial mapping between the HRV parameter and the downsampled optimal solutions $\mathcal{X}^*$ is depicted in Fig. \ref{fig:initial_mapping}. 
We assumed that, at the experiment begin, the participant was at rest ($\overline{\text{RR}}_\text{0}\!\approx\!\overline{\text{RR}}_r$) so we started from solution $\bf{h^8}$ (i.e.  $15\!-\!\frac{\sigma_r}{\Delta_{r\rightarrow s}}\!\!\approx\!8$). 
Note that the mapping depended on subject-specific features. Indeed, the range of $\overline{\text{RR}}$ was defined from the value $\overline{\text{RR}}_r$ measured in the calibration phase and the optimal solutions were associated accordingly.

The study employed a within-subjects testing design in which each participant experienced three different robot control strategies. 
Prior to the experiment, a training phase of 5 minutes was conducted to allow the user to familiarise with the task. Then, three sessions of 10 minutes were performed considering different experimental conditions. In condition \textit{(a)} and \textit{(b)}, the robot followed minimum-time and minimum-jerk trajectories, respectively, for the overall duration of the test. On the contrary, in condition \textit{(c)}, which was the core of this experimental analysis, the smoothness and total execution time of the robot trajectories were tuned online according to the human cognitive load. The order of the conditions was randomised, and there was a break between following sessions to prevent learning effects and cumulative workload. 

\subsection{Measurements and derived statistics}
In this section, we present the adopted measurements and derived statistics to assess the potentials of our framework. 
\subsubsection{Cognitive load} 
The cognitive workload induced by the collaborative task in different robot control strategies was investigated by examining the RR intervals. The latter were normalised with respect to the corresponding person-specific baseline, and statistical analysis using the non-parametric Wilcoxon signed-rank test (WSRT) was performed.

\subsubsection{Productivity}
The production rate of each operator $p$ in the various conditions was evaluated in terms of completed boxes per minute
\vspace{-0.4cm}
\begin{equation}
    \phi^{(s)}_p = \frac{60}{T^{(s)}/b^{(s)}_p},
\vspace{-0.2cm}
\end{equation}
where $b^{(s)}_p\!$ denoted the number of cycles completed in the overall duration $T^{(s)}$ of condition $s$ (i.e. 10 \si{\minute}). 

To estimate the participant's performance, an error rate statistic was defined as 
\vspace{-0.35cm}
\begin{equation}
    \epsilon^{(s)}_p = \frac{e^{(s)}_p}{b^{(s)}_p},
\vspace{-0.2cm}
\end{equation}
expressing the errors committed per cycle.
An error $e^{(s)}_p$ was marked each time the participant failed in selecting the objects with the correct code or did not place the objects in time to be picked by the robot. 

The average production and error rate statistics over all participants were used to compare the testing conditions. 

\section{EXPERIMENTAL RESULTS}
\label{sec:result}

\subsubsection{Cognitive load} 

\begin{figure}[!t]
    \centering
    \includegraphics[width=\linewidth]{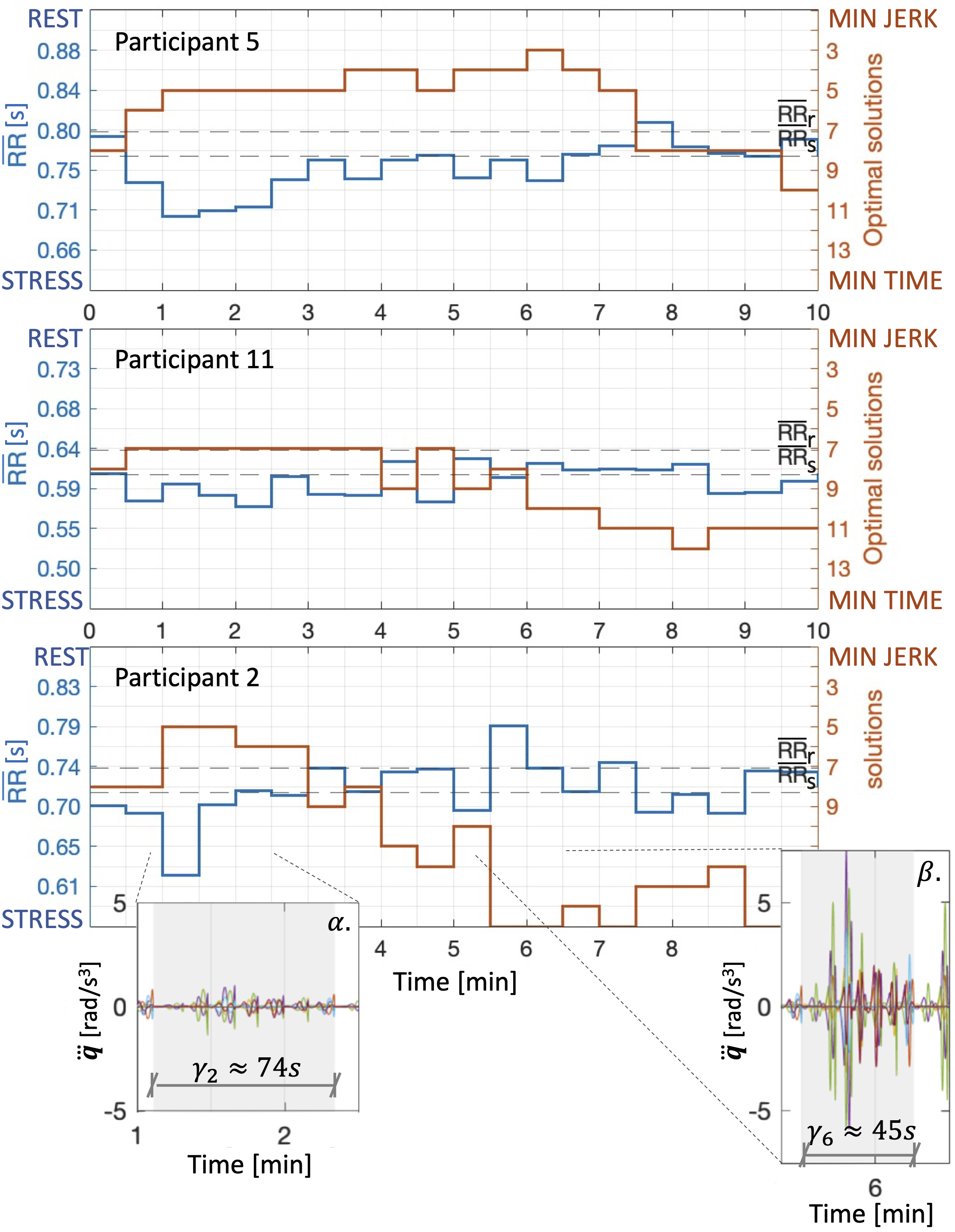}
    
    \vspace{-0.2cm}
    \caption{\small Online adaptation of trajectory smoothness and execution time as a function of detected cognitive load level for three subjects. Blue lines denote $\overline{\text{RR}}$ in time windows of 30\si{\second}. Selected optimal solution out of set $\mathcal{X}^*$ is depicted by the red line. $\alpha$. and $\beta$. highlight jerk values $\dddot{q}$ for all joints and execution time $\gamma$ of two task cycles.}
    \label{fig:CL_subjects}
\end{figure}

Figure \ref{fig:CL_subjects} shows the results of condition $c$, i.e. the online decision-making procedure based on HRV, for three out of the twelve participants. 
In the plots, the blue profile represents the mean value of RR intervals in 30-seconds time windows, while the red line depicts the selected solution. 
All participants were assumed to start from resting conditions, so the initial robot trajectory exploited the $8$-th solution of the optimal set of intervals vectors $\mathcal{X}^*$. However, it should be noticed that the associated mean RR intervals differed from one participant to the other. This was because the $\overline{\text{RR}}$ ranges were defined from the subject-specific value $\overline{\text{RR}}_r$ measured in the calibration phase. 
When a change toward a higher cognitive workload level was detected (i.e. a decrease in the blue profile), we moved toward solutions reducing the value of the kinematic quantities (joint velocity, acceleration and jerk) and thus generating smoother but slower trajectories (see subject $5$ in Fig.\ref{fig:CL_subjects}). 
Conversely, solutions resulting in fast but less smooth motions were picked when the user was relaxed (higher jerk values in close up $\beta$ of Fig.\ref{fig:CL_subjects}). 

\begin{figure}[!t]
    \centering
    \includegraphics[width=0.85\linewidth]{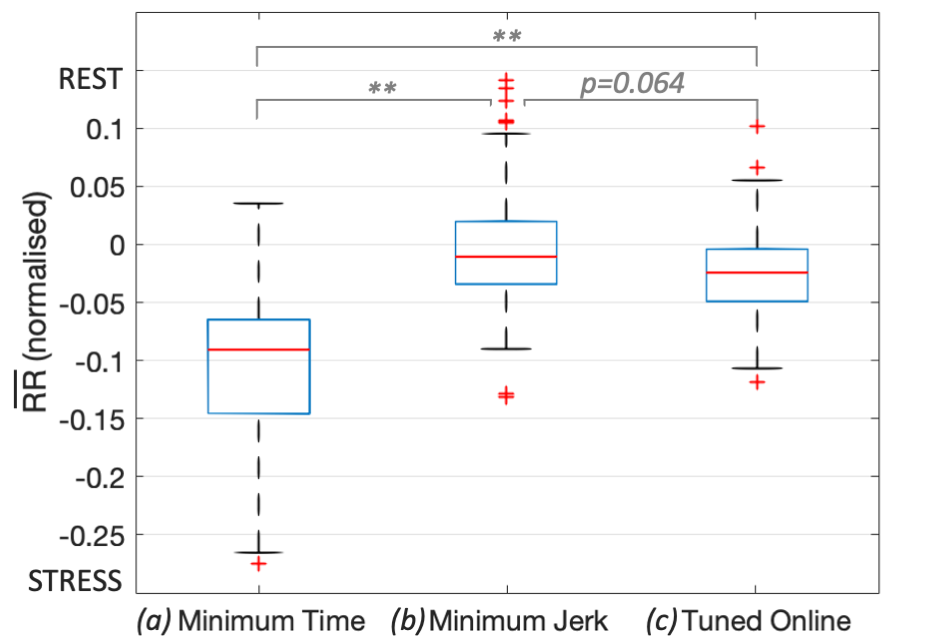}
    
    \vspace{-0.1cm}
    \caption{\small Normalised $\overline{\text{RR}}$ intervals during testing conditions.  Significance level are indicated at the *p\textless{}0.05, **p\textless{}0.005.}
    \label{fig:CL_all}
    \vspace{-0.4cm}
\end{figure}

Results of the comparison of normalised values of $\overline{\text{RR}}$ in the three different experimental conditions are depicted in Fig.\ref{fig:CL_all}. According to the employed statistic, condition $a$ was the most cognitive demanding among the tested conditions. Conversely, minimum-jerk robot movements (condition $b$) resulted in minimum levels of cognitive load and could be considered a lower bound reference for the developed workload in HRC. 
Interestingly, the cognitive demand in condition $c$  was comparable to condition $b$.
The statistical analysis consistently revealed statistically significant differences of $\overline{\text{RR}}$ in condition $a$ with respect to condition $b$ and $c$ (WSRT test, p-value$=4.88\cdot 10^{-4}$). 
On the contrary, the average $\overline{\text{RR}}$ in conditions $b$ and $c$ could not be considered statistically different (WSRT test, p-value$=0.064$). 

\subsubsection{Productivity} 
\begin{table}[!t]
    \centering
    \caption{\small Productivity statistics of different testing conditions.}
    \vspace{-0.1cm}
    \begin{tabular}{ |l||c|c|c|  }
     \hline
      & $(a)$Minimum Time & $(b)$Minimum Jerk & $(c)$Tuned Online \\
     \hline 
     $\phi$   & 1.275 & 0.300 & 1.025 \\
     $\epsilon$ & 0.537 & 0.028 & 0.196 \\
     \hline
    \end{tabular}
    \label{tab:productivity}
\vspace{-0.4cm}
\end{table}

Outcomes of the proposed productivity statistics are reported in Tab. \ref{tab:productivity}. 
According to the data acquired in this study, if the robot followed minimum-time trajectories (i.e. conditions $a$), the production rate was clearly high; however, the percentage of human errors was about 54\%. 
On the other hand, tracking minimum-jerk motions, as in conditions $b$, led to a few errors during the task accomplishment but a low production rate. 
The average productivity in the tuned condition $c$ was almost comparable to condition $a$, but the committed errors diminished considerably. 

\section{DISCUSSION}
\label{sec:discussion}

What stands out in Fig.\ref{fig:CL_subjects} is the capability of our framework to find the most appropriate solution for each human operator with his/her specific ranges of $\overline{\text{RR}}$. 
It should also be noticed that the needs of a specific subject may vary while performing the activity due to human cumulative fatigue, task demand, or even environmental factors (e.g. noise). 
Our tuning algorithm permits online adapting robot motion characteristics and pace of interaction according to the current human mental state (as registered for participant $2$ in Fig.\ref{fig:CL_subjects}).
In addition, the statistical analysis revealed that the cognitive load induced by the collaborative task in the tuned condition was comparable to the reference value of minimum mental effort experienced by the worker in HRC. This is a remarkable outcome indicating that the proposed framework is effective in mitigating the cognitive load in mixed human-robot environments. 

Concerning the human-robot throughput, increasing the robot speed always leads to a rise in the production rate. Nevertheless, it severely affects the quality of the collaborative activity. 
Our method guarantees excellent levels of average productivity without compromising human performance in terms of errors committed nor inducing mental fatigue. 

Besides, it is worth mentioning that collaborative manipulation tasks require constraint switching of different trajectories dictated by sub-task requirements (e.g. $\bf{\Dot{q}}\!=\!\bf{0}$ during object grasping), which contributes to increase the variation rate of velocity and acceleration. 
This is the reason we cannot force the velocity and smoothness of joints trajectories at the same time, and thus, we implement time/jerk minimisation.  
Further strengths of the employed multi-objective optimisation are the no need to normalise the involved objectives and the absence of an initialisation procedure, which are critical steps when considering single-objective problems with weighted terms. 

\section{CONCLUSIONS}
\label{sec:conclusion}

This paper presented an advanced HRC strategy to safeguard and potentially minimise work-related cognitive load in hybrid industrial environments.
The online adaptation of CoBot's movements to the detected human cognitive load turned out to be beneficial since participants maintained almost rest conditions while upholding high production rate. This demonstrated the potential of our framework in solving the trade-off between system productivity and human cognitive load. 
Strong points included the specificity of the framework to each operator and sensitivity to variations of load conditions during the task execution. 

In this study, the assessment of cognitive load was limited to the analysis of HRV. 
Future works could enlarge the processing to other physiological signals (i.e. galvanic skin response) for more robust monitoring, also considering physical factors. In addition, it would be interesting to include a motion capture system in the setup to track human movements during the collaboration and detect excessive work intensification and hazardous time pressure.
This would also give us the opportunity to add objectives in the trajectory optimisation relative to the worker's distance and configuration.
\vspace{-0.1cm}


\bibliographystyle{ieeetr}

\bibliography{bibliography}

\begin{thebibliography}{10}

\bibitem{Maddikunta2021}
P.~K.~R. Maddikunta, Q.-V. Pham, P.~B, N.~Deepa, K.~Dev, T.~R. Gadekallu,
  R.~Ruby, and M.~Liyanage, ``Industry 5.0: A survey on enabling technologies
  and potential applications,'' {\em Journal of Industrial Information
  Integration}, p.~100257, 2021.

\bibitem{brosque2020human}
C.~Brosque, E.~Galbally, O.~Khatib, and M.~Fischer, ``Human-robot collaboration
  in construction: Opportunities and challenges,'' in {\em 2020 International
  Congress on Human-Computer Interaction, Optimization and Robotic Applications
  (HORA)}, pp.~1--8, {IEEE}, 2020.

\bibitem{liu2020remote}
H.~Liu and L.~Wang, ``Remote human--robot collaboration: A cyber--physical
  system application for hazard manufacturing environment,'' {\em Journal of
  manufacturing systems}, pp.~24--34, 2020.

\bibitem{vanderMolene2020}
H.~F. van~der Molen, K.~Nieuwenhuijsen, M.~H.~W. Frings-Dresen, and
  G.~de~Groene, ``Work-related psychosocial risk factors for stress-related
  mental disorders: an updated systematic review and meta-analysis,'' {\em
  Occupational and environmental medicine}, 2020.

\bibitem{Dini2017}
A.~Dini, C.~Murko, S.~Yahyanejad, U.~Augsdörfer, M.~Hofbaur, and L.~Paletta,
  ``{Measurement and Prediction of Situation Awareness in Human-Robot
  Interaction based on a Framework of Probabilistic Attention},'' in {\em
  Proceedings of International Conference on Intelligent Robots and Systems
  ({IROS})}, pp.~4354--4361, {IEEE}, 2017.

\bibitem{Lagomarsino2021}
M.~Lagomarsino, M.~Lorenzini, E.~{De Momi}, and A.~Ajoudani, ``An online
  framework for cognitive load assessment in industrial tasks,'' {\em Robotics
  and Computer-Integrated Manufacturing}, vol.~78, 2022.

\bibitem{Lagomarsino2022}
M.~Lagomarsino, M.~Lorenzini, P.~Balatti, E.~D. Momi, and A.~Ajoudani, ``Pick
  the right co-worker: Online assessment of cognitive ergonomics in human-robot
  collaborative assembly,'' {\em {IEEE} Transactions on Cognitive and
  Developmental Systems}, pp.~1--1, 2022.

\bibitem{Villani2018}
V.~Villani, L.~Sabattini, C.~Secchi, and C.~Fantuzzi, ``A framework for
  affect-based natural human-robot interaction,'' in {\em Proceedings of
  International Symposium on Robot and Human Interactive Communication
  ({RO}-{MAN})}, {IEEE}, 2018.

\bibitem{Messeri2021}
C.~Messeri, G.~Masotti, A.~M. Zanchettin, and P.~Rocco, ``Human-robot
  collaboration: Optimizing stress and productivity based on game theory,''
  {\em {IEEE} Robotics and Automation Letters}, pp.~8061--8068, 2021.

\bibitem{Hopko2021}
S.~Hopko, R.~Khurana, R.~K. Mehta, and P.~R. Pagilla, ``{Effect of Cognitive
  Fatigue, Operator Sex, and Robot Assistance on Task Performance Metrics,
  Workload, and Situation Awareness in HRC},'' {\em IEEE Robotics and
  Automation Letters}, pp.~3049--56, 2021.

\bibitem{Kuehnlenz2016}
B.~Kühnlenz and K.~Kühnlenz, ``Reduction of heart rate by robot trajectory
  profiles in cooperative hri,'' in {\em Proceedings of International Symposium
  on Robotics (ISR); Conference date: 21-22 June 2016}, vol.~34, (Munich,
  Germany), pp.~400--406, {IEEE}, 2016.

\bibitem{Kuehnlenz2018}
B.~Kühnlenz, M.~Erhart, M.~Kainert, Z.-Q. Wang, J.~Wilm, and K.~Kühnlenz,
  ``Impact of trajectory profiles on user stress in close human-robot
  interaction,'' {\em Automatisierungstechnik}, pp.~483--491, 2018.

\bibitem{Flash1985}
T.~Flash and N.~Hogan, ``The coordination of arm movements: an experimentally
  confirmed mathematical model,'' {\em The Journal of Neuroscience},
  pp.~1688--1703, 1985.

\bibitem{Rojas2020}
R.~A. Rojas, E.~Wehrle, and R.~Vidoni, ``A multicriteria motion planning
  approach for combining smoothness and speed in collaborative assembly
  systems,'' {\em Applied Sciences}, vol.~10, p.~5086, 2020.

\bibitem{Boor2001}
C.~de~Boor, {\em A Practical Guide to Splines}.
\newblock Springer New York, 2001.

\bibitem{Gasparetto2007}
A.~Gasparetto and V.~Zanotto, ``A new method for smooth trajectory planning of
  robot manipulators,'' {\em Mechanism and Machine Theory}, pp.~455--471, 2007.

\bibitem{Deb2002}
K.~Deb, A.~Pratap, S.~Agarwal, and T.~Meyarivan, ``A fast and elitist
  multiobjective genetic algorithm: {NSGA}-{II},'' {\em {IEEE} Transactions on
  Evolutionary Computation}, pp.~182--197, 2002.

\bibitem{Nussinovitch2011}
U.~Nussinovitch, K.~P. Elishkevitz, K.~Katz, M.~Nussinovitch, S.~Segev,
  B.~Volovitz, and N.~Nussinovitch, ``Reliability of ultra-short {ECG} indices
  for heart rate variability,'' {\em Annals of Noninvasive Electrocardiology},
  pp.~117--122, 2011.

\end{thebibliography}

\end{document}